\documentclass{article}
\usepackage{spconf,amsmath,epsfig}
\begin{document}\sloppy

\def\x{{\mathbf x}}
\def\L{{\cal L}}
\title{Facial Expression Representation Learning by Synthesizing Expression Images}
%
\name{Kamran Ali, Charles E. Hughes}
\address{Synthetic Reality Lab, Department of Computer Science\\
University of Central Florida, Orlando, Florida\\
{\tt\small kamran@knights.ucf.edu, ceh@cs.ucf.edu}
}
\maketitle
\begin{abstract}

Representations used for Facial Expression Recognition (FER) usually contain expression information along with identity features. In this paper, we propose a novel Disentangled Expression learning-Generative Adversarial Network (DE-GAN) which combines the concept of disentangled representation learning with residue learning to explicitly disentangle facial expression representation from identity information. In this method the facial expression representation is learned by reconstructing an expression image employing an encoder-decoder based generator. Unlike previous works using only expression residual learning for facial expression recognition, our method learns the disentangled expression representation along with the expressive component recorded by the encoder of DE-GAN. In order to improve the quality of synthesized expression images and the effectiveness of the learned disentangled expression representation, expression and identity classification is performed by the discriminator of DE-GAN. Experiments performed on widely used datasets (CK+, MMI, Oulu-CASIA) show that the proposed technique produces comparable or better results than state-of-the-art methods.

\end{abstract}
\section{Introduction}
Facial expression recognition (FER) has many exciting applications in domains like human-machine interaction,  intelligent tutoring system (ITS), interactive games, and intelligent transportation. Therefore FER has been widely studied by the computer vision and machine learning community over the past several decades.  Despite this extensive research, FER is still a difficult and challenging task. Most FER techniques developed so far do not consider inter-subject variations and differences in facial attributes of individuals present in data. Hence, the representation used for the classification of expressions contains identity-related information along with facial expression information, as observed in \cite{r1} \cite{r2}. The main drawback of this entangled representation is that it negatively affects the generalization capability of FER techniques, which, as a result, degrades the performance of FER algorithms on unseen identities. In order to overcome this problem, in \cite{r36}, Bai et al. present an explicit disentanglement of facial expression features from identity representation. Similarly in \cite{r5}, Yang et al. proposes a de-expression residue learning technique to extract expression features by converting an expression image to a neutral image. 

\begin{figure}[t]
    \centering
    \includegraphics[width=10.5cm,, height=7cm]{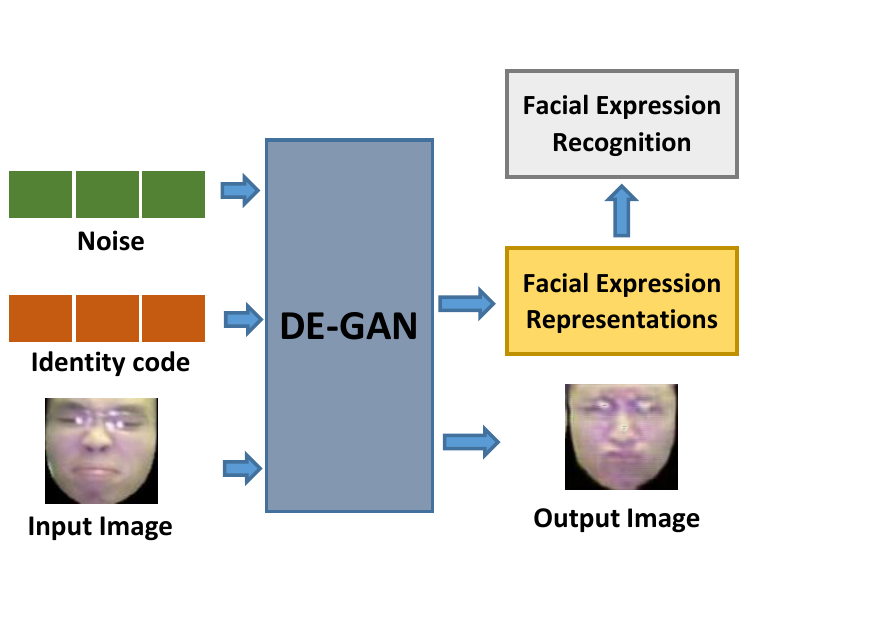}
    \caption{DE-GAN takes input image, a random noise vector and an identity code as input, and outputs disentangled and residue expressive components along with synthetic image with the same facial expression as in the input image, but with a different identity specified by the identity code. The facial expression representations are then learned for FER.}
    \label{fig:1}
\end{figure}

As shown in Figure \ref{fig:1}, we present a VAE-GAN-based expression representation learning technique that combines the concept of disentangled representation learning \cite{r34} with the effectiveness of residue learning for facial expression recognition. The proposed Disentangled Expression-learning Generative Adversarial Network (DE-GAN) is used to learn facial expressions by extracting the recorded expressive components and the disentangled expression representation through learning by synthesis procedure. The generator $G$ of DE-GAN consists of an encoder $G_{en}$ and a decoder $G_{de}$, while the discriminator $D$ of DE-GAN is a multi-task CNN. During training, the input to DE-GAN is a facial expression image $x$, an identity code $I$ and a random noise $z$, and the output of DE-GAN is a reconstructed image with the same facial expression as in $x$, but with a different identity, specified by an identity code, $I$. The noise vector $z$ is used to model factors of variations such as illumination, head pose etc. The identity code $I$ and the noise vector $z$ are then concatenated with the output of the encoder $G_{en}$ which represents the disentangled facial expression features learned by $G_{en}$. This concatenated vector is then fed to the decoder $G_{de}$ to reconstruct a facial expression image. The disentangled facial expression representation learned by the encoder $G_{en}$ is mutually exclusive of identity information, which can be best used for FER. Thus, generator, $G$, is used for two purposes: 1. to learn the disentangled facial expression representation by employing the encoder $G_{en}$, 2. to reconstruct a facial expression image, which is the output of the decoder $G_{de}$. The discriminator, $D$, of DE-GAN is trained to classify not only between real and fake images but to also perform the classification of identities and facial expressions. The estimation of facial expressions and identities in the discriminator not only helps in learning a disentangled facial expression representation, but it also improves the quality of facial expression transfer in synthetic images. 

In contrast to previous methods \cite{r36} \cite{r37} \cite{r38} \cite{r5}, which rely on residue learning to extract facial expression representation, our proposed method combines the concept of disentangled representation learning with the effectiveness of residue learning to disentangle expression representations from identity features and to improve the generalization capability of facial expression recognition. The main contributions of this paper are as follows:

\begin{itemize}
\item We present a novel disentangled and discriminative facial expression representation learning technique for FER using adversarial learning combined with residue learning in the DE-GAN framework.
\item Experimental results show that the proposed framework generalizes well to identities from various ethnic backgrounds and expression images captured both in spontaneous and posed settings.
\item The DE-GAN architecture can also be used for the tasks of facial expression transfer and editing by transferring facial expression information from input image to target identity.
\end{itemize}

\section{Related Work}
The main goal of FER is to extract features that are discriminative and invariant to variations such as pose, illumination, and identity-related information. The feature extraction process can be divided into two main categories: human-engineered features and learned features. Before the deep learning era, most of FER techniques involved human-designed features using techniques such as Histograms of Oriented Gradients (HOG) \cite{r27}, Scale Invariant Feature Transform (SIFT) features \cite{r10} and histograms of Local Phase Quantization (LPQ) \cite{r14}. 

The human-crafted features perform well in lab controlled environment where the expressions are posed by the subjects with constant illumination and stable head pose. However, these features fail on spontaneous data with varying head position and illumination. Recently, deep CNN based methods \cite{r15}, \cite{r16}, \cite{r17} \cite{r28} have been employed to increase the robustness of FER to real-world scenarios. However, the learned deep representations used for FER are often influenced by large variations in individual facial attributes such as ethnicity, gender, and age of subjects involved in training. The main drawback of this phenomenon is that it negatively affects the generalization capability of the model and, as a result, the FER accuracy is degraded on unseen subjects. Although significant progress has been made in improving the performance of FER, the challenge of mitigating the influence of inter-subject variations on FER is still an open area of research. 

Various techniques \cite{r18}\cite{r19} have been proposed in the literature to increase the discriminative property of extracted features for FER by increasing the inter-class differences and reducing intra-class variations. Most recently, Identity-Aware CNN (IACNN) \cite{r4} was proposed to enhance FER performance by reducing the effect of identity related information by using an expression-sensitive contrastive loss and an identity-sensitive contrastive loss. However, the effectiveness of contrastive loss is negatively affected by large data expansion, which is caused due to the compilation of training data in the form of image pairs \cite{r2}. Similarly, in \cite{r38} and \cite{r2}, facial expressions are transferred to a fixed identity to mitigate the effect of identity-related information. The main problem with these methods is that the accuracy of FER depends on the efficiency of the expression transfer procedure. In \cite{r5}, person-independent expression representations are learned by using residue learning. However, this technique, apart from being computationally very costly, does not explicitly disentangle the expression information from identity information, because the same intermediate representation is used to generate neutral images of the same identities. 

\section{Proposed Method}
\begin{figure}[t]
    \centering
    \includegraphics[width=9cm,, height=5.5cm]{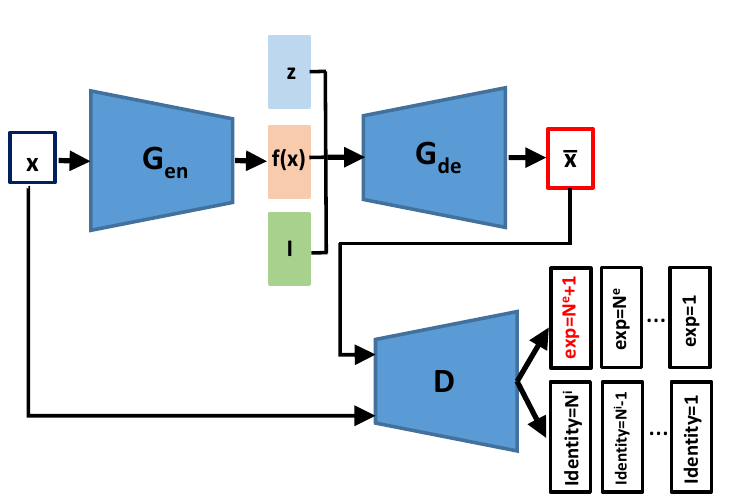}
    \caption{Architecture of our DE-GAN}
    \label{fig:2}
\end{figure}

The proposed FER technique contains two learning processes: the first is learning to disentangle an expression representation from identity information by synthesizing an expression image, and the second is learning the disentangled expression representation $f(x)$ and the expressive component from the intermediate layers of encoder $G_{en}$ of generator $G$. The overall architecture of DE-GAN is shown in Figure \ref{fig:2}. The generator $G$ of DE-GAN is based on encoder-decoder structure, while the discriminator $D$ is a multi-task CNN. The input to the encoder $G_{en}$ of $G$ is an expression image $x$ with label $y = \{y^e,y^{id}\}$, where $y^e$ denotes the expression label and the identity label is represented by $y^{id}$. The output of $G_{en}$ is a disentangled expression representation $f(x)$. This disentangled representation $f(x)$ is then concatenated with a noise vector $z$ and an identity code $I$, and fed to the decoder $G_{de}$ to synthesize an expression image $\bar{x}$ with the same expression label as $y^e$ but with different identity  $y^{id2}$, specified by identity code $I$. In order to efficiently transfer the expression of $x$ to $\bar{x}$ while preserving the identity information specified by $I$, the expression information $y^e$ must be captured in $f(x)$ in such a way that it does not contain the identity features of $x$. Thus, by explicitly feeding the identity information of $\bar{x}$ to $G_{de}$ in the form of identity code $I$, we will be able to disentangle the expression information of $x$ from its identity features in $f(x)$. Apart from producing the disentangled expression representation $f(x)$, encoder $G_{en}$ also records the expressive component from $x$ in its intermediate layers during the expression transfer process. During the second stage of learning, encoder $G_{en}$ is detached from DE-GAN after training, and the parameters of $G_{en}$ are fixed. The output of $G_{en}$, i.e  $f(x)$ is then combined with the output of intermediate layers of $G_{en}$, and fed to deep models for facial expression recognition.

\subsection{Disentanglement of Expression Representation}
The architecture of DE-GAN is based on VAE-GAN. The generator $G$ consists of an encoder $G_{en}$ and a decoder $G_{de}$, while the discriminator $D$ is a multi-task CNN. 

\subsubsection{Discriminator: }
The main objective of $D$ is three-fold: 1. to classify between real and fake images, 2. to categorize facial expressions, and 3. to recognize the identities of expression images. Therefore, discriminator $D$ is divided into two parts: $D = [D^e, D^i]$, where $D^e \in R^{N^{e+1}}$ corresponds to the part of $D$ that is used for the classification of expressions i.e $N^e$ denotes the number of expressions, and an additional dimension is used to differentiate between real and fake images. Similarly,  $D^i \in R^{N^i}$ is the part of $D$ that is used to classify the identities of expression images, where $N^i$ denotes the number of identities. The objective function of $D$ is given by the following equation:
\begin{align}
\underset{D}{\mathrm{max}}{V_D}(D,G)={}&{E_{x,y\sim p(x,y)}}[\log({D_{y^e}^e}(x)+\log({D_{y^{id}}^{i}}(x)]+\notag\\
&\underset{z\sim p_z(z),I\sim p_I(I)}{\mathrm{E_{x,y\sim p(x,y)}}}[\log({D_{N^e+1}^e}{(G(x,I,z))}]\notag\\
\end{align}
Given a real expression image $x$, the first part of the objective function of $D$ is to classify its identity and expression. The second part of the above equation shows that the objective of $D$ is also to maximize the probability of a synthetic image $\bar{x} = G_{de} (f (x),I,z)$ generated by the generator being classified as a fake class. The expression and identity classification in the discriminator $D$ helps in transferring expressions from an input image to an output image, and thus improves the disentanglement of expression features from identity information. 

\subsubsection{Generator: }
The goal of the encoder $G_{en}$ part of generator $G$ is to learn a disentangled expression representation i.e $f (x)= G_{en} (x)$ given an input expression image $x$, and to also record the expressive components of $x$ in its intermediate layers during the expression transfer process. While the decoder $G_{de}$ is used to generate a synthetic expression image $\bar{x}$: $\bar{x} = G_{de} (f (x),I,z)$, where other factor of variations like illumination, age and gender are modeled by noise vector $z$, i.e., $z \in R^{N^z}$. The identity code $I$, i.e $I \in R^N$ is in the form of a one-hot vector in which the desired identity is given by $y^{idx}$, which will be 1 in the one-hot vector. The goal of $G$ is to generate a realistic looking fake image $\bar{x}$ to fool $D$ to classify it to the identity $I$ and expression $y^e$ with the following objective function: 
\begin{align}
\underset{G}{\mathrm{max}}{V_G}(D,G) ={}&\underset{z\sim p_z(z),I\sim p_I(I)}{\mathrm{E_{x,y\sim p(x,y)}}}[\log({D_{y^e}^e}{(G(x,I,z))}+\notag\\
&\log({D_{y^{idx}}^{i}}{(G(x,I,z))}]\notag\\
\end{align}

\begin{figure}[t]
    \centering
    \includegraphics[width=9cm,, height=6cm]{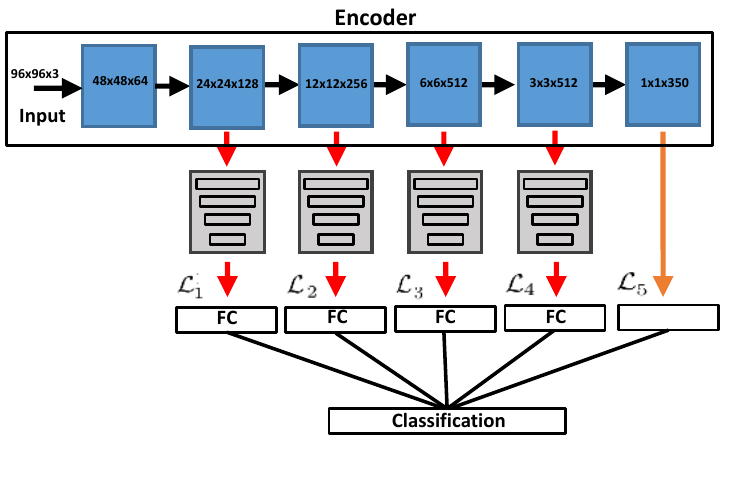}
    \caption{Second stage of learning: Facial expression recognition using expression representations. The red arrows indicate the learning of expression residue from intermediate layers of encoder, while the orange arrow indicates the learning of disentangled facial expression representation. (Best viewed in color)}
    \label{fig:3}
\end{figure}

\subsection{Facial Expression Recognition}
After training DE-GAN, the encoder $G_{en}$ is detached from DE-GAN and the parameters of $G_{en}$ are fixed. In order to learn facial expressions, the expression residue from each intermediate layer of the encoder $G_{en}$ are extracted and fed to a local CNN model. The objective function of each local CNN model is denoted by $\mathcal{L}_i$, where  $i \in [1,2,3,4]$. The disentangled expression representation $f(x)$ is then combined with the last fully connected layer of each local CNN model for FER, as shown in Figure \ref{fig:3}. The total loss function of the second stage of training is given by the following equation: 

\begin{align}
\mathcal{L}_{total}={}&\lambda_1\mathcal{L}_1 + \lambda_2\mathcal{L}_2 + \lambda_3\mathcal{L}_3+\lambda_4\mathcal{L}_4+\lambda_5\mathcal{L}_5
\label{eq:2}
\end{align}

\section{Experiments}
The proposed DE-GAN based FER technique is evaluated on three publicly available facial expression databases: CK+ \cite{r20}, Oulu-CASIA \cite{r21} and MMI \cite{r22}. 

\subsection{Implementation Details}
Face detection and face alignment is performed based on the facial landmarks detected by using Convolutional Experts Constrained Local Model (CE-CLM) \cite{r23}. Data augmentation is applied to avoid the over-fitting problem by increasing the number of training images. Therefore, five patches of size $96 \times 96$ are cropped-out from five different locations, the center and four corners of each image, and each cropped image is then rotated at ten angles i.e $-15^\circ$, $-12^\circ$, $-9^\circ$, $-6^\circ$, $-3^\circ$, $3^\circ$, $6^\circ$, $9^\circ$, $12^\circ$, $15^\circ$. Horizontal flipping is applied on each rotated image, and thus the original dataset is augmented 110 times. 

DE-GAN is initially pre-trained on the BU-4DFE ~\cite{r39} dataset, which consists of 60,600 images from 101 identities. For the optimization of the hyper-parameters the optimization strategies presented in \cite{r24} are adopted in our technique. Adam optimizer is used with a batch size of 150, learning rate of 0.0001 and momentum of 0.5. Normal distribution is used with a zero mean and standard deviation of 0.02 to initialize all network weights. DE-GAN is trained for 300 epochs, and the local FER CNN models are trained for 50 epochs. We empirically set $\lambda_1 = 0.7$, $\lambda_2 = 0.6$, $\lambda_3 = 0.4$, $\lambda_4 = 0.3$ and $\lambda_5 = 1$, and the optimal length for the disentangled expression representation is found to be 350. Contrary to conventional GAN training strategies mentioned in \cite{r3}, in later iterations of DE-GAN, when $D$ reaches to near optimal solution, $G$ is updated more frequently than $D$, due to the supervised classification provided by the class labels. 
\subsection{Experimental Results}
The \textbf{Extended Cohn-Kanade database CK+} \cite{r20} is a popular facial expression recognition database that contains 327 videos sequences from 118 subjects. Each of these sequences corresponds to one of seven expressions, i.e., anger, contempt, disgust, fear, happiness, sadness, and surprise, where each sequence starts from a neutral expression to a peak expression. To compile the training dataset, the last three frames of each sequence are extracted, which results in 981 images in total. To perform 10-fold cross validation the dataset is divided into ten different sets with no overlapping identities.

The average accuracy of 10-fold cross-validation on the CK+ database is reported in Table \ref{table:1}. It can be seen that the proposed method produces a recognition accuracy of $97.36\%$, which is higher than the accuracy of previous FER methods. Our image-based technique outperforms the sequence-based methods, where the features for FER are extracted from videos or sequences of images.

\begin{table}[t!]
\centering
\begin{tabular}{|l| c| c|} 
 \hline
 Method&Setting&Accuracy\\
 \hline
 HOG 3D\cite{r27} &Dynamic  & 91.44\\
 3DCNN \cite{r28} &Dynamic & 85.90\\
 STM-Explet\cite{r29} &Dynamic & 94.19\\
 IACNN\cite{r4} &Static  & 95.37\\
 DTAGN\cite{r30} &Static  & 97.25\\
 DeRL\cite{r5} &Static  & 97.30\\
 \hline
 CNN(baseline) &Static  & 90.34\\
 \textbf{DE-GAN(Ours)} &Static  & \textbf{97.36}\\
 \hline
\end{tabular}
\caption{CK+: 10-fold Average Accuracy for seven expressions classification.}
\label{table:1}
\end{table}

The \textbf{MMI dataset} \cite{r22} used in this experiment, contains expression images captured in a frontal view, from 31 subjects in the form of 208 video image sequences. Six basic expressions are used to label each of these sequences. The three middle frames that correspond to the peak expression are selected from each sequence to construct a dataset containing 624 images. After data augmentation, the dataset is divided into ten sets for ten-fold cross-validation, where each set has images from different identities. 

The average ten-fold cross-validation accuracy of FER on MMI database is shown in Table \ref{table:2}. The accuracy obtained using the proposed method outperforms most of state-of-the-art methods. The highest accuracy, however, is obtained by STM-Explet \cite{r29} which, as opposed to our image based method, extracts expression features from sequence of images or videos.

The \textbf{Oulu-CASIA (OC) dataset} \cite{r21} used in this experiment corresponds to the section of the OC dataset which is compiled under strong illumination condition using the VIS camera. This dataset contains 480 sequences from 80 subjects, and each sequence is labeled as one of the six basic expressions. The last three frames of each sequence are selected to create a training and testing dataset. The proposed method is evaluated on this dataset by performing a 10-fold cross-validation in which each fold is disjoint with all other folds based on subjects. 

The average 10-fold cross-validation accuracy of the proposed method on Oulu-CASIA dataset, as shown in Table \ref{table:3}, demonstrates that the proposed method outperforms all state-of-the-art techniques with average accuracy of $89.24\%$. The accuracy obtained using the proposed method is much higher than the accuracy of video based techniques.  

\begin{table}[t!]
\centering
\begin{tabular}{|l| c| c|} 
 \hline
 Method&Setting&Accuracy\\
 \hline
 HOG 3D\cite{r27}&   Dynamic  & 60.89\\
 STM-Explet\cite{r29}    &Dynamic & 75.12\\
 IACNN\cite{r4}&   Static  & 71.55\\
 DTAGN\cite{r30}&   Static  & 70.24\\
 DeRL\cite{r5}&   Static  & 73.23\\
 \hline
 CNN(baseline)& Static  & 58.46\\
 \textbf{DE-GAN(Ours)}& Static  & \textbf{72.98}\\
 \hline
\end{tabular}
\caption{MMI: 10-fold Average Accuracy for six expressions classification.}
\label{table:2}
\end{table}

\begin{table}[t!]
\centering
\begin{tabular}{|l| c| c|} 
 \hline
 Method&Setting&Accuracy\\
 \hline
 HOG 3D\cite{r27}&   Dynamic  & 70.63\\
 STM-Explet\cite{r29}    &Dynamic & 74.59\\
 Atlases\cite{r29}&   Dynamic  & 75.52\\
 PPDN\cite{r33}&   Static  & 84.59\\
 DTAGN\cite{r30}&   Static  & 81.46\\
 DeRL\cite{r5}&   Static  & 88.0\\
 \hline
 CNN(baseline)& Static  & 73.14\\
 \textbf{DE-GAN(Ours)}& Static  & \textbf{89.24}\\
 \hline
\end{tabular}
\caption{Oulu-CASIA: 10-fold Average Accuracy for six expressions classification.}
\label{table:3}
\end{table}

\section{Conclusions}
In this paper, we present a novel architecture called DE-GAN that combines the concept of disentangled representation learning with residue learning to improve the performance of FER. To achieve this goal an encoder-decoder structured generator is employed in DE-GAN, in which the disentangled expression representation is learned by transferring the expression from an input image to a synthesized image with a different identity. During the expression image synthesis process, the expression residue is also recorded in the intermediate layers of encoder. The disentangled facial expression representation is then combined with the expressive component extracted from the intermediate layers of the encoder to perform FER. The proposed method is evaluated on publicly available state-of-the-art databases, and the experimental results show that the accuracy of FER obtained by employing the proposed method is comparable or even  better than the accuracy of state-of-the-art facial expression recognition techniques. 

\bibliographystyle{IEEEbib}
\bibliography{icme2020template}

\end{document}